\documentclass{article}
\usepackage{spconf,amsmath,graphicx,booktabs,pifont,multirow}

\usepackage{amsmath}
\usepackage{amssymb}

\usepackage{threeparttable}
\usepackage{subfigure}
\usepackage{makecell}
\usepackage{pifont}
\usepackage[accsupp]{axessibility}  
\usepackage[export]{adjustbox}
\usepackage{amsfonts} 
\usepackage[rightcaption]{sidecap}
\usepackage{xcolor}
\usepackage{soul}

\title{cross branch feature fusion decoder for consistency regularization-based Semi-Supervised Change Detection
}
\name{Yan Xing$^{1}$, Qi'ao Xu$^{2}$, Jingcheng Zeng$^{2}$, Rui Huang$^{{2} {\ast}}$, Sihua Gao$^{2}$, Weifeng Xu$^{3}$, Yuxiang Zhang$^{2}$, Wei Fan$^{2}$ 
\thanks{$^{\ast}$Corresponding author: Rui Huang.}
\thanks{This work was supported in part by the Scientific Research Program of Tianjin Municipal Education Commission(2023KJ232), and the Hebei Key Laboratory of Knowledge Computing for Energy \& Power (HBKCEP202202).}
}

\address{
$^{1}$College of Safety Science and Engineering, Civil Aviation University of China, Tianjin, China \\
$^{2}$College of Computer Science and Technology, Civil Aviation University of China, Tianjin, China \\
$^{3}$Department of Computer, North China Electric Power University, Beijing, China
}

\begin{document}
%
\maketitle
%
%
\begin{abstract}
Semi-supervised change detection (SSCD) utilizes partially labeled data and a large amount of unlabeled data to detect changes. However, the transformer-based SSCD network does not perform as well as the convolution-based SSCD network due to the lack of labeled data. To overcome this limitation, we introduce a new decoder called \textit{Cross Branch Feature Fusion}~CBFF, which combines the strengths of both local convolutional branch and global transformer branch. The convolutional branch is easy to learn and can produce high-quality features with a small amount of labeled data. 
The transformer branch, on the other hand, can extract global context features but is hard to learn without a lot of labeled data. Using CBFF, we build our SSCD model based on a strong-to-weak consistency strategy. Through comprehensive experiments on WHU-CD and LEVIR-CD datasets, we have demonstrated the superiority of our method over seven state-of-the-art SSCD methods.
\end{abstract}


\section{Introduction}
\label{sec:introduction}
Semi-supervised change detection~(SSCD) aims to identify pixel-level changes occurring at the same location over different time periods by effectively utilizing a limited amount of labeled data and a large amount of unlabeled data.
It has wide applications in resource monitoring~\cite{khan2017forest,cai2021task},
disaster assessment~\cite{xu2019building},
urban management and development~\cite{liu2019semi,hafner2022urban}.

Semi-supervised methods can be classified into adversarial learning-based methods, pseudo-label-based methods, and consistency regularization-based methods. GDCNCD~\cite{gong2019generative} and SemiCDNet~\cite{peng2020semicdnet} are typical adversarial learning-based methods that use alternative optimization strategies to improve the representation learning of their respective models. Pseudo-label-based methods, RC-CD~\cite{wang2022reliable} and SemiSiROC~\cite{kondmann2023semisiroc} focus on enhancing the quality of pseudo-label and use contrast learning to improve the distinctiveness of features. Different from the above two kinds of methods, consistency regularization-based methods assume that images with strong or weak perturbs should have identical outputs ~\cite{bandara2022revisiting,mao2023semi,zhang2023semisupervised}.
Recent semi-supervised methods tend to use the consistency regularization-based framework because it is simple and has stable performance.

\begin{figure}[!t]
\graphicspath{{Fig/}}
\centering
\centerline{\includegraphics[width=\linewidth]{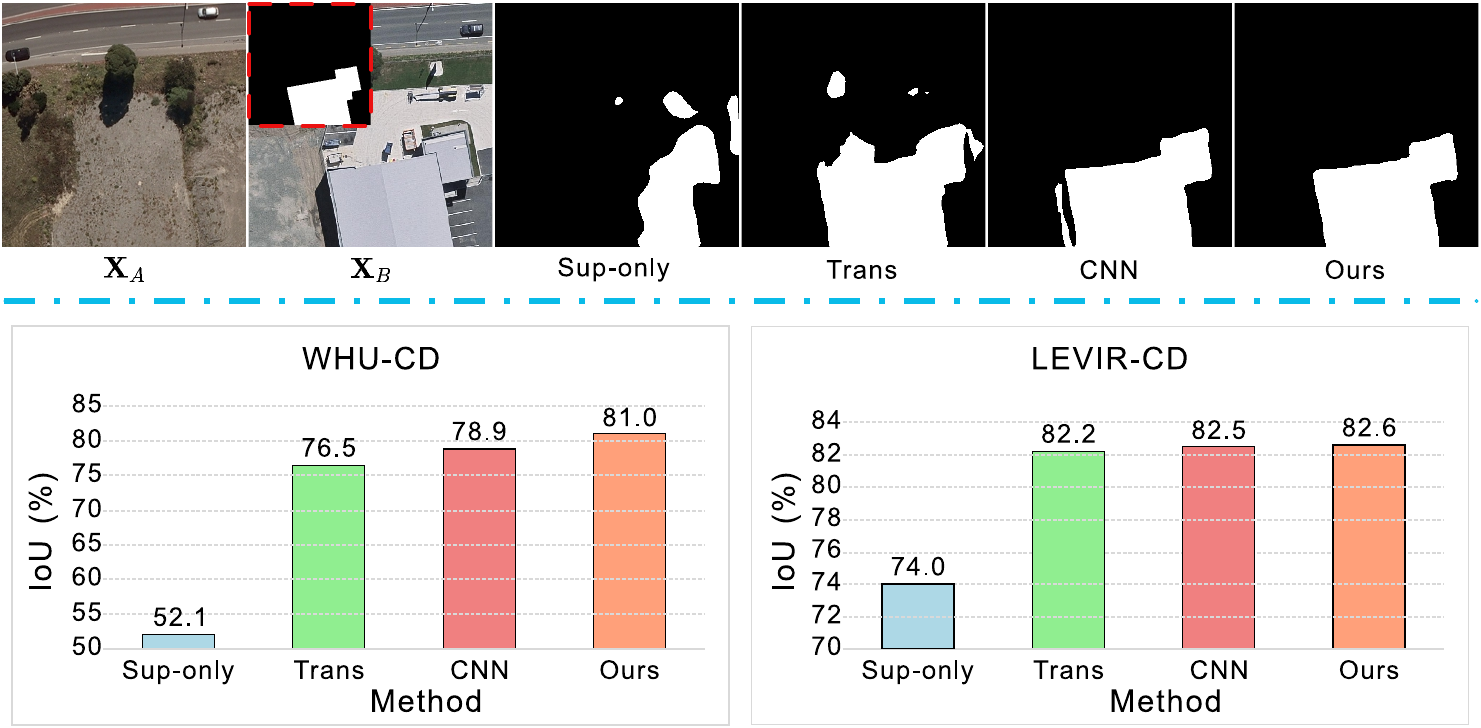}}
\caption{Motivation: Comparison of SSCD with decoders of transformer, convolution, and our proposed cross branch feature fusion by 5\% labeled training data. Sup-only denotes that our method only be trained by 5\% labeled training data.
}
\vspace{-10px}
\label{fig:motivation}
\end{figure}

The purpose of our paper is to propose a SSCD method that uses consistency regularization~\cite{sohn2020fixmatch}. Our research showed that constructing the decoder with either transformers~\cite{vaswani2017attention} or convolutional layers did not yield satisfactory results. Fig.~\ref{fig:motivation} presents the results of a UnetCD with decoder of transformer layers and convolutional layers on two public datasets~\cite{ji2018fully,chen2020spatial}. 
The model with convolution-based decoder performed better than the transformer-based model with 5\% labeled and 95\% unlabeled data. We also observed similar results in semi-supervised image classification~\cite{weng2022semi,cai2022semi}, semantic segmentation~\cite{li2022semi,huang2023semicvt}, and medical image segmentation~\cite{luo2022semi,xiao2022efficient}. We believe that transformer-based models require more high-quality labeled data, which could explain the discrepancies in performance.

\begin{figure}[!t]
\graphicspath{{Fig/}}
\centering
\centerline{\includegraphics[width=0.98\linewidth]{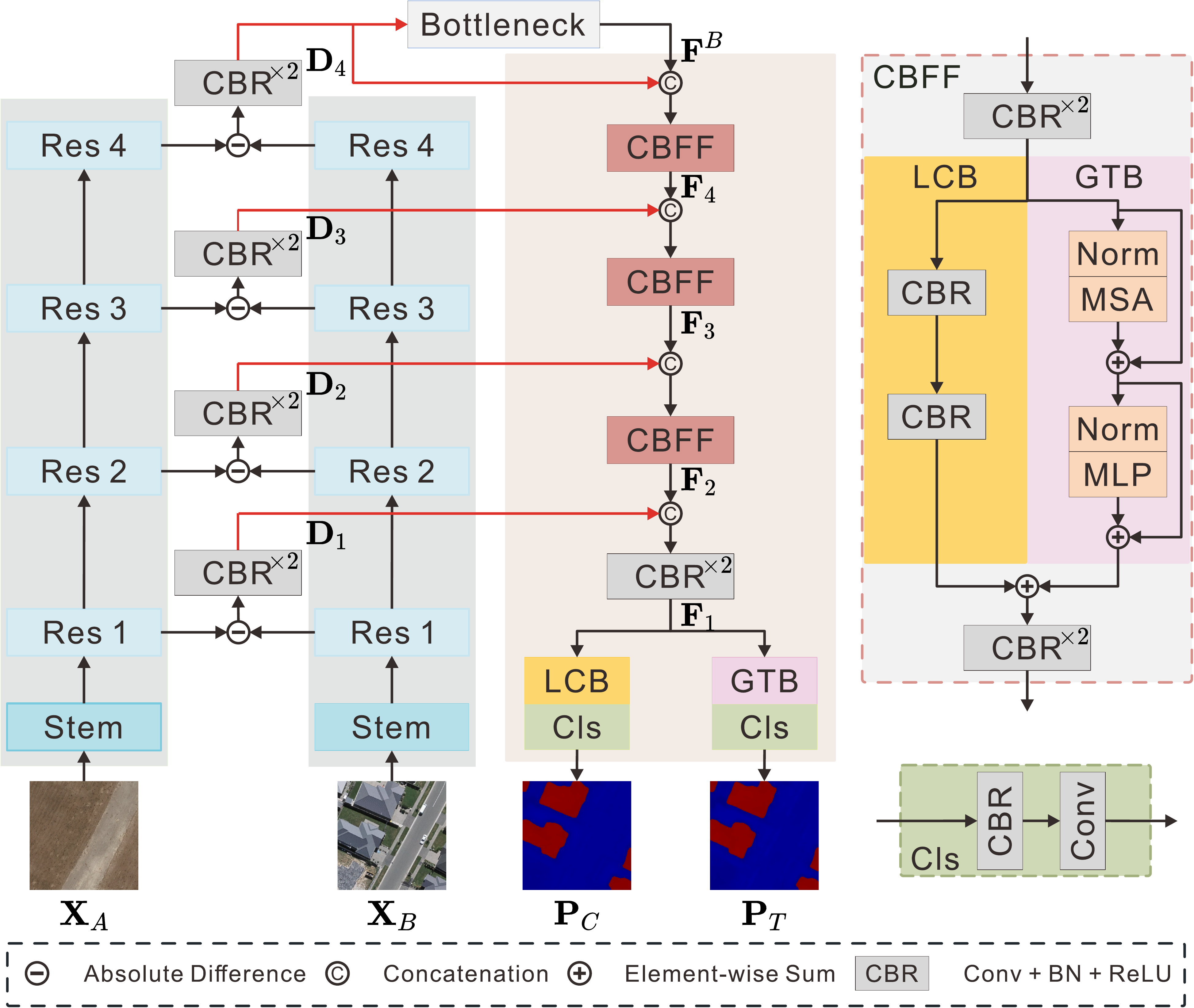}}
\caption{The architecture of our change detection network.}
\vspace{-10px}
\label{fig:framework}
\end{figure}

We propose a new decoder called \textit{Cross Branch Feature Fusion}~CBFF that effectively utilizes the features of transformer and convolution. CBFF refines features with a local convolutional branch and a global transformer branch, resulting in more representative features. The convolutional branch is easy to learn and produces high-quality features even with limited labeled data, while the transformer branch requires a lot of labeled data to learn. Our SSCD model is built using CBFF based on the strong-to-weak consistency strategy. We conduct comprehensive experiments on WHU-CD and LEVIR-CD datasets, which show that our method outperforms seven SOTA SSCD methods. The contributions of our method are as follows:
\begin{itemize}
    \item Through experimentation, we have confirmed that the convolution-based SSCD model outperforms the transformer-based SSCD model.

    \item We propose a new decoder, \textit{Cross Branch Feature Fusion}~(CBFF), that combines transformer and convolution features to enhance feature representation.
    
    \item We create an SSCD model using CBFF and consistency regularization. Numerous experiments have shown that our method is superior.
\end{itemize}


\section{Methodology}
\label{sec:propose}

\subsection{Problem formulation} 
\label{ssec:problem_statement}
Semi-supervised change detection~(SSCD) employs a limited amount of labeled data and a large amount of unlabeled data to train a change detection network to generate accurate change maps. The labeled set can be represented as $\mathcal{D}_l= \{{(\mathbf{X}^l_{Ai}, \mathbf{X}^l_{Bi}),\mathbf{Y}^l_i}\}^M_{i=1}$, where $(\mathbf{X}^l_{Ai}, \mathbf{X}^l_{Bi})$ denotes the $i$-th labeled image pair, $\mathbf{X}^l_{Ai}$ is a pre-change image, $\mathbf{X}^l_{Bi}$ is a post-change image, and $\mathbf{Y}^l_i$ is the corresponding change map. Let $\mathcal{D}_u=\{{(\mathbf{X}^u_{Ai}, \mathbf{X}^u_{Bi})}\}^N_{i=1}$ denotes the unlabeled set. 
$(\mathbf{X}^u_{Ai}, \mathbf{X}^u_{Bi})$ is the $i$-th unlabeled image pair. $M$ and $N$ indicate the number of labeled image pairs and unlabeled image pairs, respectively. In most cases, we have $N>>M$. In following sections, we will introduce the proposed change detection network, our consistency regularization-based SSCD method, and implementation details.

\subsection{Change Detection Network} 
\label{ssec:CDnetwork}
As shown in Fig.~\ref{fig:framework}, our CD network consists of a  difference feature generator, a bottleneck, three cross-branch feature fusion modules, and two prediction heads. We will give the details of each module in the following sections.

\textbf{Difference feature generator.}
The feature encoder is built on ResNet50~\cite{he2016deep} with a Siamese setup. We use the features of the first four residual modules to calculate the difference features $\mathbf{D}_i $ by
\vspace{-2px}
\begin{equation}
\vspace{-2px}
  \mathbf{D}_i = \mathrm{CBR}_3( \mathrm{CBR}_1( | \mathbf{C}_i^A - \mathbf{C}_i^B |)), i=1,2,3,4,
\end{equation}
where $\mathbf{C}_i^A$ and $\mathbf{C}_i^B$ are the features of the $i$-th residual module from image $\mathbf{X}_A$ and $\mathbf{X}_B$, respectively. $\mathrm{CBR}_k(\cdot)$ denotes a $k\times k$ convolutional layer followed with Batch Normalization and ReLU.

\textbf{Bottleneck.} 
To extract richer feature information, Atrous Spatial Pyramid Pooling (ASPP)~\cite{chen2017rethinking} is used in the bottleneck. The bottleneck feature $\mathbf{F}^B $ is calculated by 
\vspace{-2px}
\begin{equation}
\vspace{-2px}
\mathbf{F}^B = \mathrm{ASPP}( \mathbf{D}_4),
\end{equation}
where $\mathrm{ASPP}(\cdot)$ refers to the ASPP process. 

\textbf{Cross Branch Feature Fusion decoder (CBFF).}
CBFF is used to integrate the difference features and features of the previous layer. It comprises of  a Local Convolutional Branch~(LCB) and a Global Transformer Branch~(GTB). We first concatenate $\mathbf{D}_i$ and the previous layer's feature $\mathbf{F}_{i+1}$, then refine it with two convolutional operations by
\vspace{-2px}
\begin{equation}
\vspace{-2px}
\mathbf{F}_{i}^{'} = 
    \begin{cases}
    \mathrm{CBR}_3( \mathrm{CBR}_1(\mathrm{Cat}(\mathbf{D}_i, up(\mathbf{F}^B)))), & i=4,  \\
    \mathrm{CBR}_3( \mathrm{CBR}_1(\mathrm{Cat}(\mathbf{D}_i, up(\mathbf{F}_{i+1})))), & i=2,3,
    \end{cases}
\end{equation}
where $up(\cdot)$ denotes upsampling operation, $\mathrm{Cat}(\cdot,\cdot)$ is concatenate operation.

LCB makes learning easy with few labeled data using convolutional layers. The feature of LCB, $\mathbf{F}_{i}^{LCB}$, is calculated by
\vspace{-2px}
\begin{equation}
\vspace{-2px}
    \mathbf{F}_{i}^{LCB} = \mathrm{CBR}_3( \mathrm{CBR}_3(\mathbf{F}_{i}^{'})).
\end{equation}

GTB uses transformer to learn global context features. The feature of GTB, $\mathbf{F}_{i}^{GTB}$, is calculated by
\vspace{-2px}
\begin{equation}
\vspace{-2px}
\begin{aligned}
&\mathbf{Z}_i = \mathrm{MSA}(\mathrm{Norm}(\mathbf{F}_{i}^{'}))+\mathbf{F}_{i}^{'},\\
&\mathbf{F}_{i}^{GTB} = \mathrm{MLP}(\mathrm{Norm}(\mathbf{Z}_i)) + \mathbf{Z}_i,\\
\end{aligned}
\end{equation}
where $\mathrm{MLP}(\cdot)$, $\mathrm{Norm}(\cdot)$ and $\mathrm{MSA}(\cdot)$ represent multilayer perceptron, layer normalization, and multi-head self-attention, respectively.

Finally, we add the features of LCB and GTB to generate a more representative feature $\mathbf{F}_{i}$ by
\begin{equation}
    \mathbf{F}_{i} = \mathrm{CBR}_3( \mathrm{CBR}_1(\mathbf{F}_{i}^{LCB} +\mathbf{F}_{i}^{GTB})).
\end{equation}

\textbf{Change map prediction.}
To generate change maps, we first concatenate  $\mathbf{D}_1$ and the upsampled feature $\mathbf{F}_{2}$, then refine it with two convolutional operations by
\begin{equation}
    \mathbf{F}_{1} = \mathrm{CBR}_3( \mathrm{CBR}_1(\mathrm{Cat}(\mathbf{D}_1, up(\mathbf{F}_{2})))).
\end{equation}
We use two classifiers to generate change maps from the output features of LCB and GTB branches by
\begin{equation}
\begin{aligned}
        &\mathbf{P}_C = \mathrm{Cls}(\mathrm{LCB}(\mathbf{F}_{1})), \\
        &\mathbf{P}_T =\mathrm{Cls}(\mathrm{GTB}(\mathbf{F}_{1})),
\end{aligned}
\end{equation}
where $\mathrm{LCB}(\cdot)$ and $\mathrm{GTB}(\cdot)$ denote the processes of LCB and GTB, respectively. $\mathrm{Cls}(\cdot)$ consists of a $3\times3$ CBR block and a $1\times1$ convlutional layer.

\begin{figure}[!t]
\graphicspath{{Fig/}}
\centering
\centerline{\includegraphics[width=0.98\linewidth]{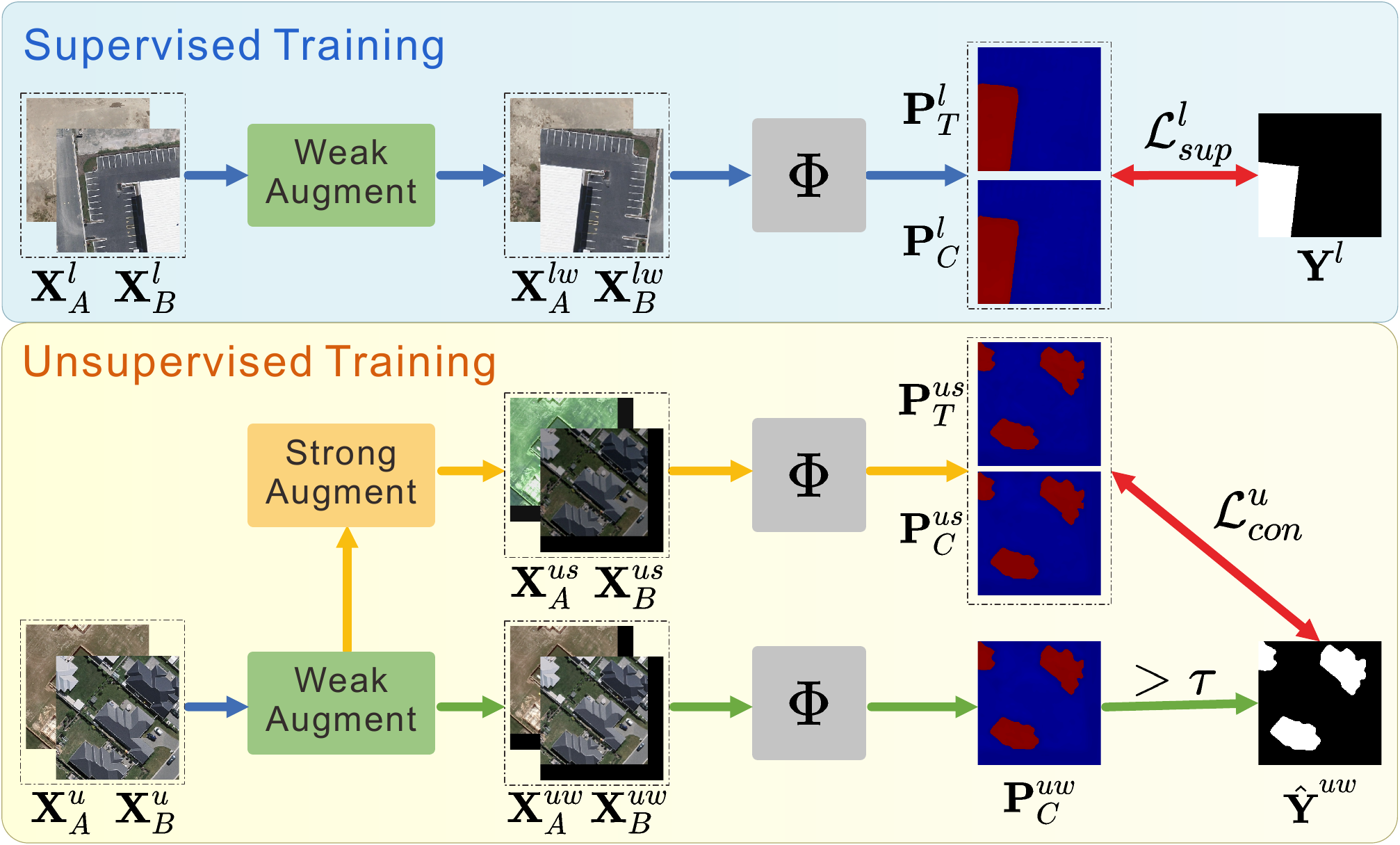}}
\caption{The framework of consistency regularization-based semi-supervised change detection method.}
\vspace{-10px}
\label{fig:semiframework}
\end{figure}

\subsection{Our consistency regularization-based SSCD method} 
\label{ssec:training_process}
Our SSCD method, shown in Fig.~\ref{fig:semiframework}, consists of supervised training part and unsupervised training part utilizing consistency regularization.

In the supervised training part, we utilize labeled dataset $\mathcal{D}_l$ to train the CD network $\Phi$. The network takes in a pair of weakly augmented images, which then generate two change maps $\mathbf{P}^l_C$ and $\mathbf{P}^l_T$. We adopt standard cross-entropy (CE) loss as supervision. Thus the loss of the supervised training part is defined as follows:
\begin{equation}
\mathcal{L}^{l}_{sup} = \frac{1}{2} ( \mathcal{L}_{CE}(\mathbf{P}^l_C,\mathbf{Y}^l)+  \mathcal{L}_{CE}(\mathbf{P}^l_T,\mathbf{Y}^l) ).
\end{equation}

In the unsupervised training part, we use a strong-to-weak consistency strategy to train $\Phi$ on the unlabeled dataset $\mathcal{D}_u$. Specifically, the output change map $\mathbf{P}^{uw}_C$ of $\Phi$ with weak augmentation input is used to generate pseudo-label $\hat{\mathbf{Y}}^{uw}$ by
\begin{equation}
\label{eq:delta}
\hat{\mathbf{Y}}^{uw} = 
    \begin{cases}
    1, & if \quad \mathbf{P}^{uw}_C > \tau  \\
    0, & else
    \end{cases}
\end{equation} 
where $\tau=0.95$ is a confidence threshold. The consistency loss of the unsupervised training part is as follows:
\vspace{-2px}
\begin{equation}
\vspace{-2px}
\mathcal{L}^{u}_{con} = \frac{1}{2} ( \mathcal{L}_{CE}(\mathbf{P}^{us}_C,\hat{\mathbf{Y}}^{uw})+\mathcal{L}_{CE}(\mathbf{P}^{us}_T,\hat{\mathbf{Y}}^{uw}) ).
\end{equation}

The total loss is composed of the supervised loss $\mathcal{L}^l_{sup}$ and the consistency loss $\mathcal{L}^u_{con}$. It can be expressed as follows:
\vspace{-2px}
\begin{equation}
\vspace{-2px}
\label{eq:total_loss}
    \mathcal{L} = \lambda_1 \mathcal{L}^l_{sup} + \lambda_2 \mathcal{L}^u_{con},
\end{equation} 
where $\lambda_1$ = 0.5 and $\lambda_2$ = 0.5.

\begin{table*}[!tb]
\centering
\caption{Quantitative comparison of different methods on WHU-CD and LEVIR-CD. The highest scores are marked in \textbf{bold}.}
\resizebox{0.93\linewidth}{!}{
\label{table:main_result}
\begin{tabular}{r|cc|cc|cc|cc|cc|cc|cc|cc}
\toprule
 \multirow{3}{*}{Method} & \multicolumn{8}{c|}{WHU-CD} & \multicolumn{8}{c}{LEVIR-CD} \\ 
  \multirow{1}{*}{} & \multicolumn{2}{c|}{5\%} & \multicolumn{2}{c|}{10\%} & \multicolumn{2}{c|}{20\%} & \multicolumn{2}{c|}{40\%} & \multicolumn{2}{c|}{5\%} & \multicolumn{2}{c|}{10\%} & \multicolumn{2}{c|}{20\%} & \multicolumn{2}{c}{40\%} \\ 
 & IoU & OA & IoU & OA & IoU & OA & IoU & OA & IoU & OA & IoU & OA & IoU & OA & IoU & OA \\
\midrule
AdvEnt~\cite{vu2019advent} & 
57.7 &	97.87 &	60.5 & 97.79 & 69.5 & 98.50 & 76.0 & 98.91 & 
67.1 & 98.15 & 70.8 & 98.38 & 74.3 & 98.59 & 75.9 & 98.67 \\ 
s4GAN~\cite{mittal2019semi} & 
57.3 & 97.94 & 58.0 & 97.81 & 67.0 & 98.41 & 74.3 & 98.85 & 
66.6 & 98.16 & 72.2 & 98.48 & 75.1 & 98.63 & 76.2 & 98.68  \\
SemiCDNet~\cite{peng2020semicdnet} & 
56.2 & 97.78 & 60.3 & 98.02 & 69.1 & 98.47 & 70.5 & 98.59 &
67.4 & 98.11 & 71.5 & 98.42 & 74.9 & 98.58 & 75.5 & 98.63  \\
SemiCD~\cite{bandara2022revisiting} & 
65.8 & 98.37 & 68.0 & 98.45 & 74.6 & 98.83 & 78.0 & 99.01 &
74.2 & 98.59 & 77.1 & 98.74 & 77.9 & 98.79 & 79.0 & 98.84  \\
RC-CD~\cite{wang2022reliable} & 
57.7 & 97.94 & 65.4 & 98.45 & 74.3 & 98.89 & 77.6 & 99.02 & 
67.9 & 98.09 & 72.3 & 98.40 & 75.6 & 98.60 & 77.2 & 98.70  \\
SemiPTCD~\cite{mao2023semi} & 
74.1 & 98.85 & 74.2 & 98.86 & 76.9 & 98.95 & 80.8 & 99.17 & 
71.2 & 98.39 & 75.9 & 98.65 & 76.6 & 98.65 & 77.2 & 98.74  \\
%
%
UniMatch~\cite{yang2023revisiting} &
78.7 & 99.11 & 79.6 & 99.11 & 81.2 & 99.18 & 83.7 & 99.29 & 
82.1 & 99.03 & 82.8 & 99.07 & 82.9 & 99.07 & 83.0 & 99.08  \\
\midrule
Ours &  
\textbf{81.0} & \textbf{99.20} 
&\textbf{81.1} & \textbf{99.18} &\textbf{83.6} & \textbf{99.29} &\textbf{86.5} & \textbf{99.43} &
\textbf{82.6} & \textbf{99.05} & \textbf{83.2} & \textbf{99.08}  &\textbf{83.2} & \textbf{99.09}  &\textbf{83.9} & \textbf{99.12}  \\

\bottomrule
\end{tabular}}
\vspace{-5px}
\end{table*}
\subsection{Implementation detail}

\textbf{Augmentation operations.} 
Weak augmentations consist of random resizing and random horizontal flipping. The resize ratio is set to a random number in $[0.8,1.2]$.  
Strong augmentations include random color jittering,  Gaussian blur, and CutMix~\cite{yun2019cutmix}. The brightness, contrast, saturation, and hue are set to $[-0.5,+0.5]$, $[-0.5,+0.5]$, $[-0.5,+0.5]$, and $[-0.25,+0.25]$, respectively. The radius of the Gaussian blur is set to a random number between 0.1 and 2.0.

\textbf{Super-parameters.}
We use PyTorch to conduct experiments and train on an NVIDIA RTX2080Ti GPU. Our model utilizes the SGD optimizer with a learning rate of 0.02, momentum of 0.9, and weight decay of 1e-4.
The total epoch is 80. And the batch size is set to 4.

\begin{figure*}[!t]
\graphicspath{{Fig/}}
\centering
\centerline{\includegraphics[width=0.92\linewidth]{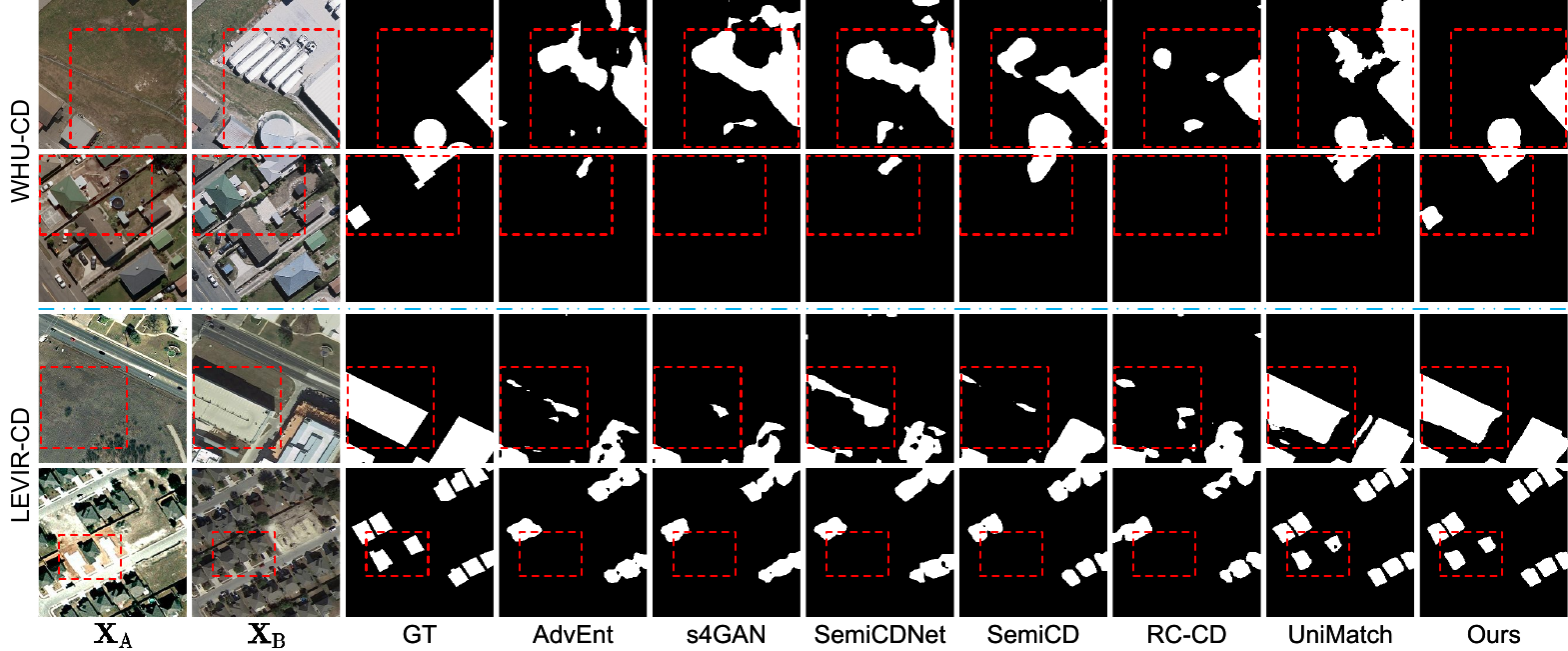}}
\vspace{-5px}
\caption{Detection results of different methods on WHU-CD and LEVIR-CD at the 5\% labeled training ratio.}
\vspace{-13px}
\label{fig:quantitative_result}
\end{figure*}
%


\section{Experiment}
\label{sec:experiment}
\subsection{Setup}
\label{ssec:setting}
{\bf Baselines.} 
We compare the proposed method with seven existing SOTA methods, including
AdvEnt~\cite{vu2019advent},
s4GAN~\cite{mittal2019semi},
SemiCDNet~\cite{peng2020semicdnet},
SemiCD~\cite{bandara2022revisiting},
RC-CD~\cite{wang2022reliable},
SemiPTCD~\cite{mao2023semi}, 
and UniMatch~\cite{yang2023revisiting}.
All methods are implemented with PyTorch and trained on the same training sets.

{\bf Datasets.}
We have conducted experiments on two widely-used benchmark datasets, namely WHU-CD~\cite{ji2018fully} and LEVIR-CD~\cite{chen2020spatial}. WHU-CD comprises two sets of aerial images, each with a resolution of \(32507\times 15354\) pixels and a pixel resolution of 0.075 m. LEVIR-CD consists of 637 high-resolution image pairs with a resolution of \(1024\times 1024\) pixels and a pixel resolution of 0.5 m. 
Following Bandara et al.~\cite{bandara2022revisiting} and Mao et al.~\cite{mao2023semi},
we crop the images into non-overlapping patches of size \(256\times 256\) and divide them into training, validation, and test sets. The training set is further divided into labeled and unlabeled data with the following ratios: \([5\%,95\%]\), \([10\%,90\%]\), \([20\%,80\%]\), \([40\%,60\%]\).

\begin{table}[ht]
\centering
\vspace{-5px}
\caption{Ablation study on the proposed decoder.}
\label{table:ablation}
\resizebox{0.95\linewidth}{!}{
\begin{tabular}{c|cc|cc|cc|cc}
\toprule
    \multirow{3}{*}{Method} & \multicolumn{4}{c|}{WHU-CD} & \multicolumn{4}{c}{LEVIR-CD}  \\
    & \multicolumn{2}{c|}{5\%} & \multicolumn{2}{c|}{10\%} & \multicolumn{2}{c|}{5\%} & \multicolumn{2}{c}{10\%}  \\ 
    & IoU & OA & IoU & OA & IoU & OA & IoU & OA \\
\midrule
    Sup-only & 52.1 & 97.24 & 57.6 & 97.84 & 74.0 & 98.53 & 78.6 & 98.82 
    \\
    CNN & 78.9 & 99.11 & 80.3 & 99.16 & 82.5 & 99.04 & 83.1 & \textbf{99.08} \\
    Trans & 76.5 & 98.97 & 80.2 & 99.13 & 82.2 & 99.03 & 83.1 & 99.07 \\
\midrule
    Ours & \textbf{81.0} & \textbf{99.20} & \textbf{81.1} & \textbf{99.18} & \textbf{82.6} & \textbf{99.05} & \textbf{83.2} & \textbf{99.08} \\
\bottomrule
\end{tabular}}
\vspace{-12px}
\end{table}

{\bf Criterion.}
Following Bandara et al.~\cite{bandara2022revisiting} and Mao et al.~\cite{mao2023semi}, we use intersection over union (IoU) and overall accuracy (OA) to evaluate different change detectors.

\subsection{Results and Discussion}
\label{ssec:results}
{\bf Comparison with the State-of-the-Art.}
Table~\ref{table:main_result} shows the quantitative comparison of different methods on WHU-CD and LEVIR-CD with different proportions of labeled data. Our method outperforms all other methods on both datasets. On WHU-CD, compared to the current SOTA method UniMatch, our method brings \(2.3\%\), \(1.5\%\), \(2.4\%\), and \(2.8\%\) performance gain in terms of IoU with \(5\%\), \(10\%\), \(20\%\), and \(40\%\) labeled data, respectively. On LEVIR-CD, the improved performance with IoU of our method over the best UniMatch are \(0.5\%\), \(0.4\%\), \(0.3\%\), and \(0.9\%\) in four partitions, respectively.


Fig.~\ref{fig:quantitative_result} shows some typical detection results of different methods on WHU-CD and LEVIR-CD under the partition of 5\%. Our approach, which incorporates both local and global information, achieves higher accuracy and more detailed results. Both quantitative and qualitative analyses support the superiority of our method.

{\bf Effectiveness of the proposed decoder.}
Table~\ref{table:ablation} displays the IoU results of various decoders to determine the effectiveness of CBFF. The CBFF-based model achieves the best performance at 5\% and 10\% partitions in both datasets. On WHU-CD, with only 5\% labeled training data, the CBFF-based model outperforms convolution-based and transformer-based models by 2.1\% and 4.5\%, respectively. These results confirm that the proposed CBFF is effective.


\section{Conclusion}
\label{sec:Conclusion}
%
In this paper, we studied semi-supervised change detection and introduced a new decoder, \textit{Cross Branch Feature Fusion}~CBFF. This decoder consists of two branches: a local convolutional branch and a global transformer branch. The convolutional branch produces high-quality features with a small amount of labeled data and is easy to learn. While the transformer branch captures global context information through multi-head self-attention. By combining the features of these two operations, CBFF generates more representative features. Using CBFF, we have built our SSCD model based on a strong-to-weak consistency strategy. We have conducted extensive experiments on WHU-CD and LEVIR-CD datasets, which demonstrate the superiority of our method over seven other state-of-the-art SSCD methods. 




\vfill
\pagebreak

\small
\bibliographystyle{IEEEbib}
\bibliography{strings,refs}
\end{document}